%% file: paper_draft.tex
\documentclass{article}
\usepackage{iclr2027_conference,times}
\usepackage{amsmath,amssymb}
\usepackage{booktabs,tabularx}
\usepackage{multirow}
\usepackage[table]{xcolor}
\definecolor{rhelmgreen}{HTML}{E7F1E8}
\definecolor{rhelmpurple}{HTML}{EEE9F5}
\definecolor{rhelmgray}{HTML}{F0F0F0}
\definecolor{rhelmlavender}{HTML}{F4EEF8}
\definecolor{rhelmlink}{HTML}{006D77}
\usepackage{graphicx}
\usepackage{wrapfig}
\usepackage{placeins}
\usepackage{algorithm}
\usepackage{algpseudocode}
\usepackage{hyperref}
\usepackage{url}
\hypersetup{
    hidelinks,
    pdftitle={MemSteer: Turning User Histories into Supervision for Agent Memory Optimization},
    pdfauthor={Jie Huang, Cuiling Lan, Qize Yu, Xiulian Peng, Yan Lu}
}
\iclrfinalcopy

\title{MemSteer: Turning User Histories into\\Supervision for Agent Memory Optimization}
\author{
    \href{https://openreview.net/profile?id=~Huangjie2}{Jie Huang}\textsuperscript{1,2}\thanks{Work done during an internship at MSRA.}\quad
    \href{https://openreview.net/profile?id=~Cuiling_Lan1}{Cuiling Lan}\textsuperscript{2}\quad
    \href{https://openreview.net/profile?id=~Qize_Yu1}{Qize Yu}\textsuperscript{1}\quad
    \href{https://openreview.net/profile?id=~Xiulian_Peng1}{Xiulian Peng}\textsuperscript{2}\quad
    \href{https://openreview.net/profile?id=~Yan_Lu7}{Yan Lu}\textsuperscript{2}\\[0.4em]
    \normalfont\textsuperscript{1}\href{https://www.pku.edu.cn}{Peking University}\quad
    \textsuperscript{2}\href{https://www.microsoft.com}{Microsoft Research Asia}
}

\begin{document}
\maketitle
\fancyhead{}
\fancyhead[L]{Preprint}
\input{abstract}
\input{introduction}
\input{related_work}
\input{method}

\input{experiments}
\input{conclusion}
\input{statements}

\clearpage

\input{paper_draft.bbl}
\clearpage
\appendix
\input{appendix}
\end{document}

%% file: abstract.tex
\begin{abstract}
An effective agent memory system should adapt to a user's history and task requirements.
Human-annotated supervision can guide this adaptation by providing feedback
for evaluating and improving memory designs.
Yet suitable labels are rarely available for new user histories,
and manually constructing QA pairs and supporting evidence is costly.
We introduce MemSteer, which turns unlabeled user histories into question--answer
(QA) supervision with supporting evidence for memory-design optimization.
MemSteer constructs an evidence-grounded answer plan that specifies how facts
across related records jointly support a reference answer.
The plan guides question generation, while source-grounded auditing and blind
answer recovery check the resulting annotations.
To study the utility of this supervision, we develop MemSteer-Opt,
which combines QA-based design evaluation with evidence-guided failure diagnosis
to revise memory store, retrieval, and context construction without updating
model parameters.
Experiments on ATM-Bench and RHELM show that this supervision is competitive
with official-QA-guided optimization.
Relative to Base-BM25, MemSteer-Opt improves ATM-Bench Question Score by 13.4 points
and RHELM accuracy by 4.1 and 6.6 points without and with external data, respectively.
The supervision also benefits two existing optimizers on ATM-Bench,
while new-user experiments demonstrate continued design optimization using
supervision generated from the target histories.
\end{abstract}

%% file: introduction.tex
\section{Introduction}

\begin{wrapfigure}[21]{r}{0.48\textwidth}
    \vspace{-52pt}
    \centering
    \setlength{\abovecaptionskip}{4pt}
    \includegraphics[width=\linewidth]{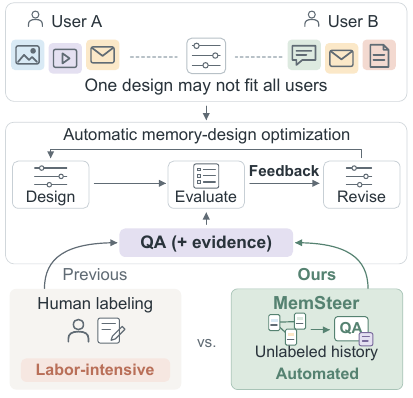}
    \caption{Motivation for MemSteer.
    User histories vary in data characteristics, so one memory design may not suit every user.
    Automatic design adaptation through iterative optimization requires feedback on memory performance.
    MemSteer supports this optimization by automatically generating QA--evidence supervision from unlabeled histories, avoiding labor-intensive manual labeling.}
    \label{fig:motivation}
\end{wrapfigure}
Agents that assist users over time need to draw on information accumulated
across conversations, emails, and documents.
An external memory system supports this process by storing and organizing
the user's history and retrieving relevant context for the agent.
Storage granularity, retrieval strategy, and answer-context construction
shape how effectively the agent can use that history.
User histories differ in their content, modalities, and organization,
so a memory design effective for one user may be less suitable for another.
This motivates adapting memory designs to user histories.

Recent methods such as MemEvolve, ALMA, and $M^{\star}$ optimize memory designs
using downstream task feedback~\citep{zhang2025memevolve,xiong2026alma,pan2026mstar}.
While executable tasks can provide feedback through their outcomes, personal
records and interactions often lack an explicit signal for assessing memory quality.
Annotated question--answer (QA) pairs can supply this signal~\citep{pan2026mstar},
but constructing questions, reference answers, and supporting evidence for each
new history is costly.
This motivates our question (Figure~\ref{fig:motivation}):
\emph{Can unlabeled user histories guide memory-design optimization?}

Generating QA from the history itself offers a potential source of this feedback.
Within agent memory, MemBuilder uses synthetic QA rewards to train
memory-construction policies~\citep{shen2026membuilder}, while MemMA uses QA probes
to repair stored memories~\citep{lin2026memma}.
We instead investigate whether history-derived QA can provide reliable feedback
for evaluating and improving memory-system designs.
For memory-design optimization, reference answers determine how candidate designs
are evaluated, while source evidence can help diagnose unsuccessful responses.
Unsupported answers or mismatched evidence can misdirect design selection
and later revisions.

We propose MemSteer, a framework that turns unlabeled user histories
into source-grounded QA--evidence supervision for memory-design optimization
(Figure~\ref{fig:method}).
It brings together related source records and specifies how their facts
jointly support a reference answer. The resulting evidence-grounded answer plan
guides question generation. Source-grounded auditing checks
the generated QA for evidence support and consistency, while blind answer
recovery assesses whether the reference answer can be recovered from
the sampled records.

To investigate the utility of MemSteer's generated supervision,
we develop MemSteer-Opt, an evidence-guided memory-design optimizer.
It uses source evidence and execution traces to identify whether
incorrect responses stem from storage, retrieval, or answer generation,
then proposes targeted design revisions and evaluates them using QA performance.
Starting from a basic memory system, we use MemSteer-Opt to optimize
memory representations, retrieval strategies, and context construction.
We further supply the generated QA to M$^{\star}$ and MemEvolve
\citep{pan2026mstar,zhang2025memevolve},
allowing each framework to optimize within its own design space.

Our contributions are threefold:
\begin{list}{\labelitemi}{\setlength{\leftmargin}{1.25em}\setlength{\labelwidth}{0.75em}\setlength{\labelsep}{0.5em}}
    \item \textbf{Memory-design optimization from unlabeled histories.}
    MemSteer constructs QA--evidence supervision from unlabeled user histories
    for evaluating and improving memory designs without updating model parameters.
    \item \textbf{Evidence-grounded answer planning and verification.}
    We construct answer plans that make explicit how source facts jointly
    support a reference answer for question generation.
    Source-grounded verification checks the resulting annotations.
    \item \textbf{Empirical validation across optimizers and user histories.}
    Experiments with MemSteer-Opt on ATM-Bench and RHELM
    \citep{mei2026atmbench,zhang2026rhelm} show that generated
    supervision improves memory designs and is competitive with the
    Official-QA reference. Further experiments demonstrate its utility for
    two existing optimizers on ATM-Bench and for continued optimization
    on new user histories.
\end{list}

%% file: related_work.tex
\section{Related Work}

\paragraph{Memory-design optimization.}
Applications differ in what agents need to remember and how they use
past information, making memory design an application-specific
engineering challenge~\citep{xiong2026alma,pan2026mstar}.
Recent work therefore explores automatic memory-design optimization
to accommodate these requirements while reducing reliance on manual design.
ALMA searches over code-based memory designs using task success
and interaction logs as feedback~\citep{xiong2026alma}.
MemEvolve introduces a dual-level process that evolves both
experience memory and its underlying architecture~\citep{zhang2025memevolve}.
$M^{\star}$ searches over memory programs by jointly refining their
schema, logic, and instructions through task feedback and reflective
code evolution~\citep{pan2026mstar}.
At a narrower level, CluE refines memory-extraction prompts
using downstream feedback across heterogeneous tasks~\citep{yang2026clue}.
These methods optimize memory designs using feedback from downstream tasks.
In many personal-memory applications, however, user records and everyday
interactions do not naturally yield success-or-failure signals
to guide memory-design optimization.
\emph{We therefore investigate how these histories can be turned into supervision
for memory-design optimization.}

\paragraph{Synthetic queries and QA for training and evaluation.}
Outside agent memory, synthetic data has been used to reduce
annotation requirements for model training and system evaluation.
Promptagator generates query--document pairs to train task-specific
dense retrievers~\citep{dai2022promptagator}.
\citet{alberti2019roundtrip} generate QA by selecting answers,
formulating questions, and filtering pairs through answer-recovery
consistency before training reading-comprehension models.
For structured synthesis, GraphGen samples knowledge-graph subgraphs
to construct atomic, aggregated, and multi-hop QA for supervised
fine-tuning~\citep{chen2025graphgen}.
For evaluation, DataMorgana constructs corpus-specific RAG benchmarks
with configurable user and question categories~\citep{filice2025datamorgana},
while Ragas uses graph relationships to synthesize multi-hop
test questions~\citep{ragas_test_generation}.
The usefulness of synthetic questions has also been studied directly:
\citet{yue2022qve} learn a question value estimator to select examples
that improve target-domain QA performance, rather than relying only
on generation or answer-model scores.
\emph{We ask whether synthetic QA can provide effective feedback
for memory-design optimization.}

\paragraph{Synthetic supervision for agent memory.}
Generated QA has also been incorporated into memory learning
and refinement.
MemBuilder and CMI-Mem generate QA pairs from current sessions
and retrieved memories, then use answering accuracy as a reward signal
to train LLMs to construct and update memories via reinforcement
learning~\citep{shen2026membuilder,wang2026cmimem}.
MemMA generates QA probes from each session and its historical context,
then uses answering failures to repair the memory bank~\citep{lin2026memma}.
MemMA also examines how the choice of probe generator affects repair outcomes.
MemSteer instead uses generated QA pairs with source-grounded evidence
to guide memory-design optimization, rather than to train memory models
or repair stored memories.
This turns user histories into a source of feedback for memory-design
optimization without requiring manually annotated QA for the target history.

%% file: method.tex
\begin{figure}[!t]
    \centering
    \includegraphics[width=\linewidth]{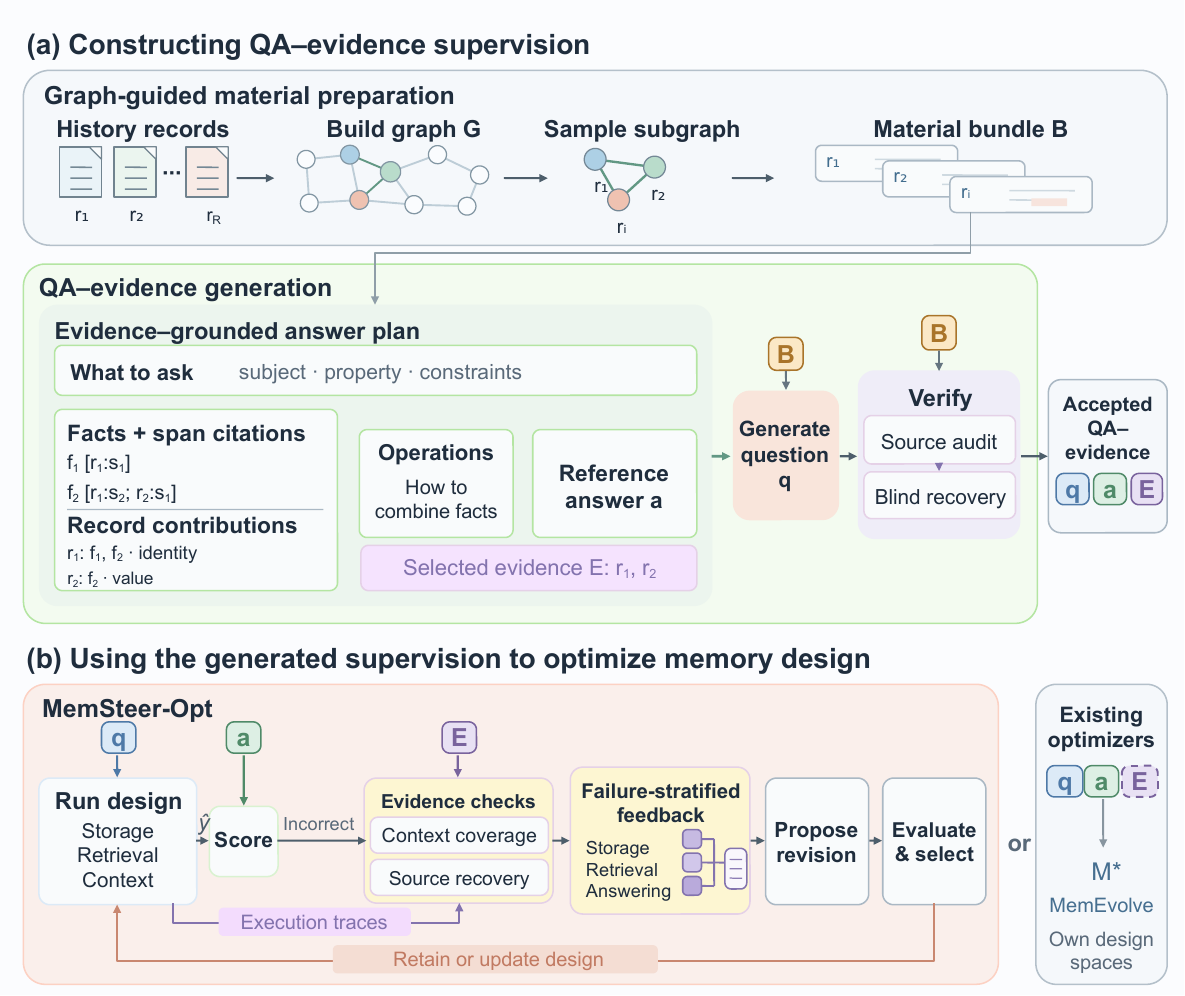}
    \caption[Overview of MemSteer.]{Overview of MemSteer.
    (a) MemSteer turns unlabeled user histories into QA--evidence supervision
    through evidence-grounded answer planning, question generation,
    and source-grounded verification.
    (b) The resulting supervision supports the evaluation and refinement
    of memory designs with either MemSteer-Opt, our evidence-guided optimizer,
    or existing memory-design optimizers.
    Dashed $E$ denotes optional evidence input.}
    \label{fig:method}
\end{figure}

\section{Method}

\subsection{Overview}

Histories from different users vary in content and structure,
so a memory design effective for one user may be less suitable for another.
Given unlabeled user histories $H$, our goal is to improve agent
memory designs without manually annotating QA pairs for those histories.
To this end, MemSteer constructs QA--evidence supervision:
\begin{equation}
    \mathcal{D}_{H}=\{(q_i,a_i,E_i)\}_{i=1}^{N},
\end{equation}
where $q_i$ is a question about the history,
$a_i$ is its reference answer,
and $E_i$ identifies the selected source records, with source-span citations
providing provenance for the supporting facts.
Reference answers provide a criterion for evaluating memory designs,
while evidence annotations support the diagnosis of unsuccessful responses.

We describe the supervision construction process
in Section~\ref{sec:construction}.
To assess the effectiveness of this supervision for
memory-design optimization, we develop MemSteer-Opt,
which evaluates candidate designs and uses evidence-guided feedback
to propose iterative revisions (Section~\ref{sec:evolution}).
The generated supervision can also be supplied to existing optimizers,
allowing each to retain its own design space and optimization procedure.

\subsection{Constructing QA--evidence supervision}
\label{sec:construction}

MemSteer constructs QA--evidence annotations by selecting related source records,
establishing how their facts support an answer,
and generating and verifying the corresponding question.

\paragraph{Graph-guided material preparation.}
We organize each user's history as a graph of source records, preserving original
text, identifiers, and source-provided structural metadata. Edges connect records
sharing a conversation, occupying adjacent document sections, explicitly
referencing one another, or having sufficient lexical overlap.
Path-based expansion from a seed samples a connected subgraph, whose records form
a context-budgeted material bundle $B$ with source-span references for identifying
jointly supporting facts and evidence records.

\paragraph{Evidence-grounded answer planning and question generation.}
Given a material bundle $B$, an LLM-based planner constructs
an evidence-grounded answer plan that specifies what the question should ask
and its reference answer $a$, identifies supporting facts with source-span
citations, and describes each selected evidence record's contribution
and operations needed to combine facts.

Answer plans that fail programmatic checks of structural consistency
and provenance coverage are rejected before question generation.
Given the answer plan and material bundle $B$,
a question generator produces a standalone question.

\paragraph{Source-grounded verification.}
We verify each candidate through a semantic audit followed by blind
answer recovery.
Given the material bundle $B$, the answer plan, and the generated
question, an LLM reviewer returns structured verification judgments
with supporting rationales.
These judgments assess answer support and question--plan
consistency, with separate assessments of each fact's source support
and each evidence record's stated contribution.
A candidate proceeds only if all required judgments are supported.

Given only the question, its answer type, and the source records in $B$,
an LLM-based blind reader returns an answer with source citations.
A candidate passes only if the reader declares the question answerable,
its answer exactly matches the reference after type-specific normalization,
and its citations cover every selected evidence record.
Normalization rules are detailed in Appendix~\ref{app:exp-metrics}.

We partition the generated QA--evidence annotations into disjoint
training and validation sets, using the former to generate revision proposals
and the latter to select memory designs.

\subsection{MemSteer-Opt: evidence-guided memory-design optimization}
\label{sec:evolution}

MemSteer-Opt uses the generated QA--evidence supervision to evaluate
memory designs and turn observed failures into feedback for iterative revisions.

\paragraph{Design space and evaluation.}
Starting from a basic memory system, MemSteer-Opt iteratively optimizes
how historical records are organized into memory units and indexed,
how relevant records are retrieved,
and how retrieved records are assembled into the context supplied to the answerer.
We represent these storage, retrieval, and context-construction choices
as a memory design $d=(s,r,c)$.

Under design $d$, the system constructs a memory $M_d$
from the user's historical records $H$:
\[
 M_d=\operatorname{Store}_{s}(H).
\]
For each question $q_i$, it retrieves information from $M_d$,
assembles the answer context $X_{d,i}$,
and generates a response $\hat a_{d,i}$:
\begin{equation}
 X_{d,i}=\operatorname{Context}_{c}
       (\operatorname{Retrieve}_{r}(q_i,M_d)),\qquad
 \hat a_{d,i}=\operatorname{Answer}(q_i,X_{d,i}).
\end{equation}
We evaluate each prediction $\hat a_{d,i}$ against its reference answer $a_i$
under a fixed scoring protocol, yielding a binary correctness score
$y_{d,i}\in\{0,1\}$.
We score a design $d$ by its accuracy on the same validation set as
$A(d)=\frac{1}{N_{\mathrm{val}}}\sum_{i=1}^{N_{\mathrm{val}}}y_{d,i}$,
where $N_{\mathrm{val}}$ is the number of validation questions.
We use this binary validation accuracy for design selection; final benchmark
scores follow the corresponding evaluation protocols (Appendix~\ref{app:exp-metrics}).
We hold the answering model, its instruction prompt,
and the final context budget fixed across designs.

\paragraph{Evidence-guided diagnosis and sampling.}
Beyond scoring candidate designs, we use the automatically constructed
\emph{evidence annotations} to build targeted feedback for revising memory designs.

\emph{(i) Evidence-guided diagnosis.}
For each incorrect training-set response, we use its evidence annotations
to check whether the required source records and cited passages
are fully available in the final answer context.
If so, we label an \emph{answering failure}.
For missing evidence, we use the annotated source IDs to check whether the
current memory system can recover the corresponding records.
We assign a \emph{storage failure} if any missing record is unrecoverable,
and a \emph{retrieval failure} if all missing records are recoverable.

\emph{(ii) Failure-stratified sampling.}
To cover different failure modes within the proposer's input-context budget,
we sample cases separately from each observed failure category.
Each sampled case includes the question, its reference answer and annotated
source evidence, together with the current system's response and diagnostic results.
Together with the current design and revision history,
these cases form the input to the design proposer.

\paragraph{Feedback-guided revision and selection.}
Given this feedback, an LLM-based design proposer proposes a revised design $d'$,
using the diagnosed failures to guide changes to storage, retrieval,
or context construction.
It also provides a rationale for the proposed changes.
We rebuild the memory under each valid candidate and evaluate its accuracy
on the fixed validation set.
We replace the current design $d_t$ with $d'$ only if $A(d')>A(d_t)$;
otherwise, we retain $d_t$.
After accepting a revision, we rerun the updated system on the training set
to refresh the failure diagnoses and sampled feedback for the next iteration.

\paragraph{Supervision for existing memory-design optimizers.}
Beyond MemSteer-Opt, we use the generated supervision to guide existing
memory-design optimizers within their own design spaces.
Generated questions and reference answers supply task feedback for evaluating
and refining memory designs; source evidence can additionally support feedback
when the optimizer's interface accommodates it.
This tests whether the supervision remains useful across optimization frameworks.

%% file: experiments.tex
\section{Experiments}

We address four questions:
\textbf{Q1} (\S\ref{sec:exp-q1}): Can generated supervision guide memory-design optimization?
\textbf{Q2} (\S\ref{sec:exp-transfer}): Can optimized memory designs transfer and adapt?
\textbf{Q3} (\S\ref{sec:exp-q2}): Which supervision-construction choices matter?
\textbf{Q4} (\S\ref{sec:exp-feedback}): Does evidence-guided feedback improve optimization?

\subsection{Experimental Setup}
\label{sec:exp-setup}

\providecommand{\gainentry}[2]{\begin{tabular}[t]{@{}c@{}}#1\\[-3pt]{\tiny\color{green!60!black}$\uparrow$#2}\end{tabular}}
\providecommand{\gainentrybold}[2]{\begin{tabular}[t]{@{}c@{}}\textbf{#1}\\[-3pt]{\tiny\color{green!60!black}$\uparrow$\textbf{#2}}\end{tabular}}
\providecommand{\lossentry}[2]{\begin{tabular}[t]{@{}c@{}}#1\\[-3pt]{\tiny\color{red!70!black}$\downarrow$#2}\end{tabular}}

\paragraph{Benchmarks and evaluation.}
\textbf{ATM-Bench}~\citep{mei2026atmbench} is a multimodal personal-memory QA benchmark
built from approximately four years of personal records. Its corpus contains 11,034
email, image, and video records. Unless otherwise specified, we follow the
ATM-Bench scoring protocol and evaluate performance on its 1,013-question main
set and 31-question hard multi-hop set.
\textbf{RHELM}~\citep{zhang2026rhelm} covers 21,142 interaction records from ten
personas, evaluated with and without external data sources.
Benchmark selection is discussed in Appendix~\ref{app:benchmark-selection}.
\paragraph{Optimizers and protocol.}

MemSteer-Opt starts from a basic record-level retrieve-and-read system,
which we refer to as Base-BM25: it indexes
source text and metadata and uses the original question for BM25~\citep{robertson2009bm25}
retrieval. Retrieved content is assembled in ranked order within a fixed context
budget and passed to a fixed answering model.
$M^\star$~\citep{pan2026mstar} and MemEvolve~\citep{zhang2025memevolve} retain their
own starting systems and design spaces.

MemSteer uses our generated QA--evidence supervision for both training and validation while official QA is used only for testing.
MemSteer-Opt runs 20 reflective evolution iterations per setting.

\paragraph{Models and metrics.}
On ATM-Bench, DeepSeek-V4-Flash~\citep{deepseekai2026v4} serves as the answerer,
open-answer judge~\citep{zheng2023judge}, and design proposer, matching the
answerer and judge used by the corresponding leaderboard
configurations~\citep{atmbench2026leaderboard}.

We report Question Score (QS) and evidence recall at 10 (R@10).

On RHELM, we follow the official protocol, using
GPT-4.1-mini~\citep{openai2025gpt41} for answering and evaluation and reporting
overall accuracy, separately with and without external data.
We include existing memory systems as performance references; their configurations
and additional results are detailed in Appendix~\ref{app:reference-systems} and Appendix~\ref{app:exp-rhelm-full}.

\subsection{Can Generated Supervision Guide Memory-Design Optimization?}
\label{sec:exp-q1}

\begin{table}[!htbp]
\centering
\begin{minipage}[t]{0.50\textwidth}
\vspace{0pt}
\centering
\footnotesize
\setlength{\tabcolsep}{2pt}
\setlength{\abovecaptionskip}{0pt}
\setlength{\belowcaptionskip}{3pt}
\renewcommand{\arraystretch}{1.00}
\caption{ATM-Bench results across memory systems and supervision sources (\%).
``+'' denotes optimization from each optimizer's start; arrows show point changes.
Bold scores mark the best per metric, excluding Oracle. Oracle uses gold evidence.
$^\dagger$Official-QA training items are included in evaluation.}
\label{tab:q1-atm}
\begin{tabular*}{\linewidth}{@{\extracolsep{\fill}}l cc|cc@{}}
\toprule
 & \multicolumn{2}{c|}{\textbf{Full (1,013)}} & \multicolumn{2}{c}{\textbf{Hard (31)}} \\
\cmidrule(lr){2-3} \cmidrule(lr){4-5}
\textbf{Method} & \textbf{QS} & \textbf{R@10} & \textbf{QS} & \textbf{R@10} \\
\midrule
\multicolumn{5}{@{}l}{\textit{Existing memory systems (DeepSeek-V4)}} \\
Memexa & 65.3 & 78.9 & -- & -- \\
A-Mem & -- & -- & 16.7 & 38.5 \\
MemPalace & -- & -- & 13.6 & 28.9 \\
HippoRAG2 & -- & -- & 12.3 & 32.2 \\
\midrule
\rowcolor{rhelmgray}
Oracle {\scriptsize (gold evidence)} & 71.8 & -- & 46.2 & -- \\
\midrule
\multicolumn{5}{@{}l}{\textit{Our optimizer: MemSteer-Opt}} \\
Base-BM25 & 54.9 & 61.7 & 8.2 & 10.0 \\
\quad + Official-QA$^\dagger$ & 67.3 & 83.5 & 27.9 & 42.2 \\
\rowcolor{rhelmgreen}
\quad + \textbf{MemSteer} & \gainentrybold{68.3}{13.4} & \gainentrybold{84.1}{22.4} & \gainentrybold{29.1}{20.9} & \gainentrybold{43.6}{33.6} \\
\midrule
\multicolumn{5}{@{}l}{\textit{External optimizers}} \\
$M^\star$ start & 47.4 & 55.2 & 1.9 & 13.6 \\
\quad + Official-QA$^\dagger$ & 51.8 & 69.4 & 9.5 & 26.0 \\
\rowcolor{rhelmgreen}
\quad + \textbf{MemSteer} & \gainentry{53.4}{6.0} & \gainentry{69.8}{14.6} & \gainentry{12.9}{11.0} & \gainentry{25.6}{12.0} \\
\addlinespace[2pt]
MemEvolve start & 35.1 & 41.3 & 7.1 & 18.7 \\
\quad + Official-QA$^\dagger$ & 53.7 & 63.8 & 11.4 & 21.6 \\
\rowcolor{rhelmgreen}
\quad + \textbf{MemSteer} & \gainentry{53.7}{18.6} & \gainentry{63.7}{22.4} & \gainentry{9.5}{2.3} & \lossentry{18.6}{0.1} \\
\bottomrule
\end{tabular*}
\end{minipage}\hfill
\begin{minipage}[t]{0.46\textwidth}
\vspace{0pt}
\centering
\centering
\footnotesize
\setlength{\tabcolsep}{1.5pt}
\setlength{\abovecaptionskip}{0pt}
\setlength{\belowcaptionskip}{3pt}
\renewcommand{\arraystretch}{1.00}
\caption{RHELM results (accuracy, \%). w/o: conversation records only;
w/: conversation records plus emails and documents.
We use GPT-4.1-mini in MemSteer-Opt and other optimized memory systems.}
\label{tab:q1-rhelm}
\begin{tabular*}{\linewidth}{@{\extracolsep{\fill}}l cc@{}}
\toprule
\textbf{Method} & \textbf{w/o} & \textbf{w/} \\
\midrule
\multicolumn{3}{@{}l}{\textit{Existing methods (GPT-4.1-mini)}} \\
RAG ($k{=}20$) & 23.5 & 25.1 \\
RAG ($k{=}50$) & \textbf{28.9} & 32.6 \\
Hybrid ($k{=}5$) & 16.7 & 17.6 \\
Hybrid ($k{=}20$) & 24.8 & 26.0 \\
Hybrid ($k{=}50$) & 27.8 & 29.6 \\
MemGPT & 13.9 & 17.3 \\
Mem0 & 21.1 & 28.9 \\
MemU & 23.1 & 33.6 \\
\midrule
\multicolumn{3}{@{}l}{\textit{Long-Context Models (1M)}} \\
Gemini-2.5-Flash-Lite & 16.0 & 17.3 \\
Qwen-2.5-14B-Instruct & 15.3 & 8.1 \\
GPT-4.1-mini & 24.0 & 33.9 \\
\midrule
\multicolumn{3}{@{}l}{\textit{Our optimizer: MemSteer-Opt}} \\
Base-BM25 & 22.3 & 33.3 \\
\quad + Official-QA$^\dagger$ & 26.5 & \textbf{40.7} \\

\rowcolor{rhelmgreen}
\quad + \textbf{MemSteer} & \gainentry{26.4}{4.1} & \gainentry{39.9}{6.6} \\
\midrule
\multicolumn{3}{@{}l}{\textit{Transferred ATM-Bench design}} \\
ATM-Bench-evolved(frozen) & 26.4 & 38.6 \\
\quad + Official-QA$^\dagger$ & \textbf{31.5} & \textbf{42.9} \\

\rowcolor{rhelmgreen}
\quad + \textbf{MemSteer} & \gainentry{30.6}{4.2} & \gainentry{42.8}{4.2} \\

\bottomrule
\end{tabular*}
\end{minipage}
\end{table}

Tables~\ref{tab:q1-atm} and~\ref{tab:q1-rhelm} compare MemSteer-generated
supervision with official annotations (Official-QA) on ATM-Bench and RHELM;
the ATM comparison also includes two external optimizers. Official-QA uses
600 of the 1,013 official ATM items and 788 of the 1,305 RHELM items for training
and validation. These items remain in its evaluation sets, whereas MemSteer
uses no evaluation items for training. The final RHELM MemSteer configuration
uses 788 generated training items and is evaluated on all 1,305 official QA items.

Based on Tables~\ref{tab:q1-atm} and~\ref{tab:q1-rhelm}, we make the following observations.

\noindent\textbf{Generated supervision improves memory designs.}
With MemSteer supervision, MemSteer-Opt raises QS from 54.9 to 68.3 on
ATM-Bench and from 8.2 to 29.1 on Hard, alongside gains in evidence recall
(Table~\ref{tab:q1-atm}). On RHELM, MemSteer improves over Base-BM25 by
4.1 accuracy points without external data and 6.6 points with it
(Table~\ref{tab:q1-rhelm}).

\noindent\textbf{Generated supervision is competitive with the human-annotated reference.}
On ATM-Bench, MemSteer scores marginally higher than Official-QA across all four metrics,
including QS margins of 1.1 points on the main set and 1.3 points on Hard.
On RHELM, MemSteer remains 0.1 and 0.8 accuracy points below
Official-QA without and with external data, respectively.

\noindent\textbf{The supervision also benefits existing optimizers.}
From their own starting systems, $M^\star$ and MemEvolve gain 6.0 and 18.6
main-set QS points with MemSteer supervision, compared with 4.4 and 18.6
points with Official-QA. Thus, the supervision's utility is not limited to
MemSteer-Opt. The benefit is not uniform: on Hard, MemEvolve obtains smaller
gains than with Official-QA, and its R@10 decreases by 0.1 points.

\noindent\textbf{The optimized designs are competitive with existing memory systems.}
On ATM-Bench, MemSteer-Opt achieves the highest QS among the non-oracle systems
using DeepSeek-V4 as the answerer. On RHELM with external data, MemSteer
reaches 39.9 accuracy, compared with 33.9 for the strongest listed RAG,
long-context, or memory-system baseline. Without external data, it remains
competitive with the strongest listed baseline, RAG with $k{=}50$.

\par\WFclear
\subsection{Can Optimized Memory Designs Transfer and Adapt?}
\label{sec:exp-transfer}

We test design reuse across benchmarks and continued
optimization for users, respectively.

\noindent\textbf{Cross-benchmark transfer.}
We freeze the design optimized with generated ATM-Bench QA and apply it to
RHELM records without target-domain optimization.
ATM-Bench-evolved(frozen) outperforms Base-BM25 by 4.1 and 5.3 accuracy points
without and with external data, respectively (Table~\ref{tab:q1-rhelm}).
These results suggest that some of the design improvements learned on ATM-Bench
remain useful on RHELM without target-domain optimization.
When further optimized using either RHELM Official-QA or QA generated from RHELM
histories, the transferred design achieves further gains. Generated supervision
achieves performance close to the Official-QA reference without requiring manually
annotated target-domain QA for optimization.

\begin{wraptable}{r}{0.52\textwidth}
\centering
\footnotesize
\setlength{\abovecaptionskip}{0pt}
\setlength{\belowcaptionskip}{3pt}
\setlength{\tabcolsep}{1.5pt}
\renewcommand{\arraystretch}{1.06}
\caption{RHELM cross-user accuracy (\%, with external data).
Martin and David are outside the six-user pool; personal continuation adapts its
shared design using target-history-generated supervision. Source-user scores
pool evaluation questions.}
\label{tab:exp-transfer}
\label{tab:q3}
\begin{tabular*}{\linewidth}{@{\extracolsep{\fill}}lcc|c@{}}
\toprule
\shortstack{\textbf{Evaluation}\\\textbf{users}} & \textbf{Base-BM25} & \shortstack{\textbf{6-user}\\\textbf{design}} & \shortstack{\textbf{Personal}\\\textbf{continuation}} \\
\midrule
Source users (6) & 31.5 & 34.3 & -- \\
\midrule
Martin Keller & 41.2 & 37.7 & \textbf{44.3} \\
David R.\ Ellis & 37.3 & 38.6 & \textbf{39.2} \\
\bottomrule
\end{tabular*}
\vspace{\dimexpr-\intextsep+3pt\relax}
\end{wraptable}
\noindent\textbf{Cross-user transfer and continuation.}
We apply a frozen design optimized on six RHELM users to Martin Keller and
David R.\ Ellis, who are excluded from the source-user pool. We then continue
optimization from this shared design using QA--evidence supervision generated
from each target user's history, reserving official QA for evaluation
(Appendix~\ref{app:exp-benchmarks}).

Direct transfer helps David slightly but reduces Martin's accuracy
(Table~\ref{tab:exp-transfer}). Personal continuation improves both transferred
designs, especially for Martin, suggesting that target-history supervision can
support adaptation beyond a shared design.

\subsection{What Makes QA--Evidence Supervision Effective?}
\label{sec:exp-q2}

We first compare QA-generation methods and the effect of verification on
supervision quality and downstream utility. We then examine two generator
design choices: constructing an explicit evidence-grounded answer plan and
separating its construction from question generation.

\begin{wraptable}{R}{0.56\textwidth}
\vspace{-\intextsep}
\centering
\footnotesize
\setlength{\tabcolsep}{1.5pt}
\setlength{\abovecaptionskip}{0pt}
\setlength{\belowcaptionskip}{3pt}
\renewcommand{\arraystretch}{1.00}
\caption{QA-generation methods and verification on ATM-Bench.
QA quality scores are reported for two
judges (DS-V4F: DeepSeek-V4-Flash); Avg. is their mean.
QS measures downstream utility; Kept is the percentage of candidates
retained after verification (``--'': no filtering).}
\label{tab:q2}
\begin{tabular*}{\linewidth}{@{\extracolsep{\fill}}l cc|ccc|c@{}}
\toprule
 & \multicolumn{2}{c|}{\textbf{QS}} & \multicolumn{3}{c|}{\textbf{QA quality (\%)}} & \textbf{Kept} \\
\cmidrule(lr){2-3} \cmidrule(lr){4-6}
\textbf{Method} & \textbf{Full} & \textbf{Hard} & \shortstack{\textbf{GPT-5-}\\\textbf{mini}} & \textbf{DS-V4F} & \textbf{Avg.} & \textbf{(\%)} \\
\midrule
Naive-QA & 54.2 & 5.3 & 63.3 & 63.8 & 63.5 & -- \\
\quad + checks & 64.8 & 24.4 & 77.8 & 77.0 & 77.4 & 21.2 \\
Ragas-QA & 63.0 & 23.3 & 73.5 & 69.3 & 71.4 & -- \\
\quad + checks & 65.1 & 20.7 & 84.3 & 80.8 & 82.5 & 42.1 \\
\shortstack[l]{MemSteer\\w/o checks} & 65.2 & 22.0 & 83.5 & 67.3 & 75.4 & -- \\
\rowcolor{rhelmgreen}
\textbf{MemSteer} & \textbf{68.3} & \textbf{29.1} & \textbf{94.3} & \textbf{87.5} & \textbf{90.9} & \textbf{47.9} \\
\bottomrule
\end{tabular*}
\vspace{\dimexpr-\intextsep+3pt\relax}
\end{wraptable}

\noindent\textbf{QA-generation ablations.}
Table~\ref{tab:q2} compares heuristic synthesis that randomly selects records
to form a material bundle (Naive-QA), Ragas multi-hop
generation~\citep{ragas_test_generation} (Ragas-QA), and MemSteer, each with and
without verification. Checks denote our source-grounded verification
(\S\ref{sec:construction}).

For this comparison, QA quality is assessed using four criteria: premise support,
question clarity, answer support, and evidence support. For each judge, a question
receives 1 only when all four criteria are supported, and 0 otherwise.
The QA quality score is the mean of these binary
scores, expressed as a percentage. We evaluate QA quality using two independent
LLMs---GPT-5-mini~\citep{openai_gpt5mini} and DeepSeek-V4-Flash---and their average
to see whether the trends are consistent.

We make the following observations.
\textbf{(i)} MemSteer outperforms Naive-QA and Ragas-QA, even when both
baselines use our checks, in downstream QS and in QA quality assessed by both judges.
\textbf{(ii)} Our checks improve QA quality across all three generation methods
and downstream QS in most settings, except Ragas-QA on Hard. Naive-QA's
lowest retention indicates that a larger fraction of its candidates fail
the verification criteria.

\textbf{(iii)} QA quality scores from GPT-5-mini and DeepSeek-V4-Flash
show broadly consistent trends across generation methods and generally align
with downstream QS. This supports using the QA quality score as a useful
indicator of supervision quality.

\begin{wraptable}{R}{0.53\textwidth}
\vspace{-\intextsep}
\centering
\footnotesize
\setlength{\abovecaptionskip}{0pt}
\setlength{\tabcolsep}{1.7pt}
\renewcommand{\arraystretch}{1.1}
\caption{Generated QA quality evaluation. q denotes a question.}
\label{tab:qa-generation-design}
\begin{tabular*}{\linewidth}{@{\extracolsep{\fill}}lrr@{}}
\toprule
\textbf{Generation} & \textbf{Quality (\%)} & \shortstack{\textbf{Tokens/q}\\\textbf{(k)}} \\
\midrule
Direct QA & 67.8 & 64.0 \\
Joint Plan--Q & 73.1 & 68.2 \\
\rowcolor{rhelmgreen}
\textbf{Two-stage Plan--Q} & \textbf{85.9} & 105.1 \\
\bottomrule
\end{tabular*}
\vspace{\dimexpr-\intextsep+3pt\relax}
\end{wraptable}

\noindent\textbf{Evidence-grounded answer planning and staged generation.}
We test whether constructing an evidence-grounded answer plan improves QA quality
over direct generation, and whether programmatically checking the plan before a
separate question-generation call provides further gains over joint generation.
Table~\ref{tab:qa-generation-design} compares three generation procedures on the
same 600 tasks, holding source records, question types, evidence-count requirements,
and the DeepSeek-V4-Flash generator fixed. \textit{Direct QA} jointly generates
the question, answer, and evidence in one call, without an intermediate
answer plan. \textit{Joint Plan--Q} first produces an evidence-grounded answer plan
and then the question within the same response.
\textit{Two-stage Plan--Q} constructs the answer plan, applies programmatic
structural and provenance checks, and generates the question in a separate
call using the accepted answer plan and source records.

All three use the same GPT-4.1-mini filter, which sees the QA, evidence and source
material but not our detailed evidence-grounded answer plan.
This shared filter is used only for the
generation comparison, rather than the full MemSteer verification pipeline.
For QA passing this filter, a separate, blinded GPT-4.1 review assesses QA quality
with four criteria as used in Table~\ref{tab:q2}.

Joint Plan--Q improves QA quality over Direct QA, although both use one call,
supporting explicit evidence-grounded answer planning. Two-stage Plan--Q improves
quality further with programmatic plan checks before separate question generation,
but has the highest per-question token cost.

\noindent\textbf{Training-question budget.}
Increasing the budget from 100 to 200 questions yields the largest performance
gains in the evaluated sweep. Beyond 200, retrieval recall largely plateaus,
while QS improves overall on both ATM-Bench and ATM-Bench-Hard
(Appendix~\ref{app:atm-training-budget}).

\subsection{Does Evidence-Guided Feedback Improve Optimization?}
\label{sec:exp-feedback}

\begin{figure*}[!htbp]
\centering
\includegraphics[width=\textwidth]{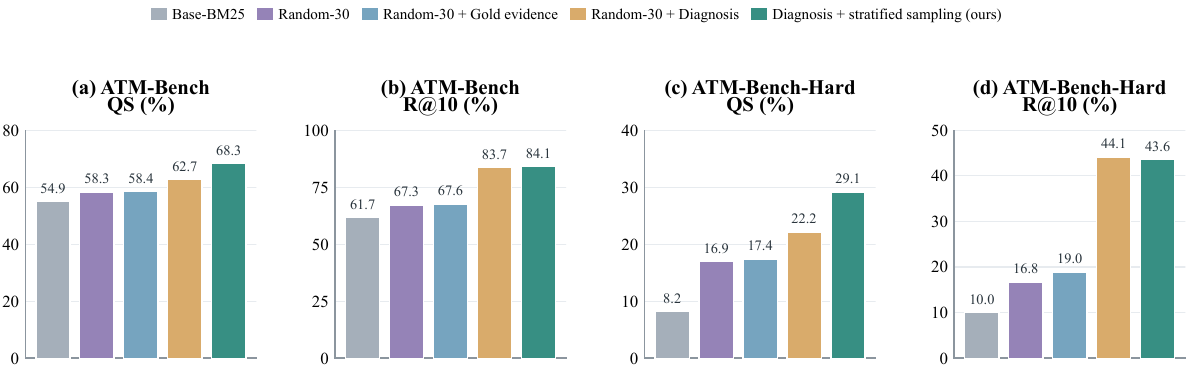}
\caption{Attribution and sampling ablation on ATM-Bench and
ATM-Bench-Hard. Bars report QS and R@10 (\%). Optimized conditions use
MemSteer-generated supervision and start from Base-BM25;
\textit{Diagnosis + stratified sampling} is the full MemSteer-Opt configuration.}
\label{fig:exp-feedback}
\end{figure*}

We test whether explicit failure diagnosis improves on raw-evidence feedback
and whether failure-stratified sampling further improves memory-design optimization.
All four feedback conditions use MemSteer-generated QA--evidence supervision
and start from Base-BM25 (Figure~\ref{fig:exp-feedback}).

We compare four ways to construct feedback from unsuccessful cases.
\textbf{Random-30} randomly samples up to 30 scored failures per round and gives
the design proposer their questions, reference answers, predictions, scores, retrieval
traces, and answer-context excerpts. \textbf{Random-30 + Gold evidence} retains
the random sampling and adds required-evidence IDs and gold-record excerpts,
but no failure attribution. \textbf{Random-30 + Diagnosis} further adds
evidence-grounded failure attribution while retaining random sampling.
\textbf{Diagnosis + stratified sampling (ours)} keeps the diagnostic feedback
and samples up to ten cases per failure category, preventing frequent failure
types from exhausting the feedback budget.
Appendix~\ref{app:exp-feedback-inputs} details the feedback fields, while
Appendix~\ref{app:exp-trajectories} shows an accepted architectural trajectory.

We make the following observations. \textbf{(i)} Adding gold evidence provides only
limited improvement over trace-based feedback. \textbf{(ii)} Explicit failure
attribution yields larger gains, particularly in evidence retrieval, supporting
the use of stage-level diagnostic feedback rather than raw evidence alone.
\textbf{(iii)} Combining diagnosis with stratified sampling achieves the highest
downstream QS, although diagnosis with random sampling yields higher evidence
recall on Hard.

%% file: conclusion.tex
\section{Conclusion}

We presented MemSteer, a framework for constructing QA--evidence supervision
from unlabeled user histories and using it to guide memory-design optimization.
Rather than requiring manually annotated QA for each new history, MemSteer
derives reference answers and supporting evidence from the records themselves.
These annotations serve complementary purposes: reference answers support the
evaluation of candidate memory designs, while source evidence helps diagnose
unsuccessful responses and guide revisions. We instantiate this approach in
MemSteer-Opt and also use the generated supervision with existing optimizers
that retain their own design spaces. Experiments on ATM-Bench and RHELM show
that generated supervision improves memory designs, achieves performance
competitive with the Official-QA reference, and supports continued adaptation
to new user histories.

%% file: statements.tex
\section*{Ethics Statement}
Our experiments use released benchmark data under their published terms, without
collecting new personal data or conducting human-subjects studies. Source records
are processed through third-party LLM APIs, and generated QA may reveal sensitive
information by combining facts across records. Deployment on personal histories
should therefore require informed user consent and appropriate privacy safeguards.
MemSteer is intended to improve memory systems for users, not to enable
unauthorized profiling.

\section*{Reproducibility Statement}
Section~\ref{sec:construction} specifies the supervision-construction pipeline
(graph-guided material preparation, evidence-grounded answer planning, question
generation, and source-grounded verification), and
Section~\ref{sec:evolution} specifies the design space, evaluation protocol,
evidence-guided diagnosis, and acceptance rule of MemSteer-Opt.
Section~\ref{sec:exp-setup} describes the benchmarks, baselines, iteration
budget, and model roles.
Appendix~\ref{app:exp-details} documents additional experimental details, including
evaluation populations and feedback inputs; Appendix~\ref{app:exp-more-results}
reports further experimental results.
Because the pipeline depends on remote LLM APIs, exact outputs may vary with
model nondeterminism and provider updates; we therefore report the model
identities used for each role.

\section*{LLM Usage}
LLMs perform evidence-grounded answer planning, question generation,
source-grounded auditing, and blind answer recovery in MemSteer. They also
propose design revisions in MemSteer-Opt, act as fixed answerers and answer
judges, and perform shared filtering and post hoc QA-quality evaluation.
We use DeepSeek-V4-Flash, GPT-4.1-mini, GPT-4.1, and GPT-5-mini across these
roles (Section~\ref{sec:exp-setup}).
In preparing this paper, we also used LLM assistants for language editing and
coding assistance.
The authors reviewed all such output and take full responsibility for this paper.

%% file: appendix.tex
\begingroup
\raggedbottom
\setlength{\textfloatsep}{10pt plus 2pt minus 2pt}
\setlength{\floatsep}{8pt plus 2pt minus 2pt}
\setlength{\intextsep}{8pt plus 2pt minus 2pt}
\setlength{\abovecaptionskip}{6pt}
\setlength{\belowcaptionskip}{4pt}
\setlength{\parskip}{4pt}
\section{Additional Experimental Details}
\label{app:exp-details}

\subsection{Benchmark Selection}
\label{app:benchmark-selection}

We prioritize ATM-Bench~\citep{mei2026atmbench} and
RHELM~\citep{zhang2026rhelm} because recent work characterizes
performance on LoCoMo~\citep{maharana2024locomo} and
LongMemEval~\citep{wu2024longmemeval} as near-saturated in
conversational-memory QA settings~\citep{he2026memoryarena}.
ATM-Bench combines real-world images, videos, and emails, whereas
RHELM couples evolving simulated user dialogues with external textual
sources such as emails and documents.

\subsection{Reference Systems}
\label{app:reference-systems}

Table~\ref{tab:q1-atm} compares DeepSeek-V4-based Memexa~\citep{memexa2026},
A-Mem~\citep{xu2025amem}, MemPalace~\citep{mempalace2026}, and
HippoRAG2~\citep{gutierrez2025hipporag2} on ATM-Bench~\citep{atmbench2026leaderboard}.

Table~\ref{tab:q1-rhelm} includes the RHELM reference results~\citep{zhang2026rhelm}
for RAG ($k\in\{20,50\}$), Hybrid ($k\in\{5,20,50\}$),
MemGPT~\citep{packer2023memgpt}, Mem0~\citep{chhikara2025mem0}, and
MemU~\citep{nevamind2025memu}, all using GPT-4.1-mini as the answering model.
Hybrid combines dense retrieval with BM25 through reciprocal rank fusion.
The same table reports 1M-context inference with
Gemini-2.5-Flash-Lite~\citep{comanici2025gemini25},
Qwen-2.5-14B-Instruct (1M)~\citep{yang2025qwen251m}, and
GPT-4.1-mini~\citep{openai2025gpt41}.
The Base-BM25, optimized, and transferred-design rows follow the protocols
in Sections~\ref{sec:exp-setup} and~\ref{sec:exp-transfer}.

Appendix Table~\ref{tab:rhelm-full} additionally reports RAG with
GPT-4.1~\citep{openai2025gpt41}, Gemini-2.5-Pro~\citep{comanici2025gemini25},
and Claude-Opus-4.5~\citep{anthropic2025opus45}, all at $k=20$.

Native Qwen3-VL/GPT-5-mini results, including
ATM-RAG~\citep{mei2026atmbench}, Self-RAG~\citep{asai2023selfrag},
ScrapMem~\citep{chang2026scrapmem}, and MemoryOS~\citep{kang2025memoryos}, are
included in Appendix Table~\ref{tab:atm-combined-appendix}.

\subsection{Evaluation populations}
\label{app:exp-benchmarks}

The training split drives failure attribution and revision proposals, whereas
validation is used to accept revisions and select the final design. MemSteer
uses generated QA--evidence supervision for these processes, without using
benchmark evaluation questions. The Official-QA reference instead uses official
QA for training and validation; these items remain included in evaluation.

On ATM-Bench, both the Official-QA and MemSteer conditions use 400 training items
and a fixed 200-item validation set; the former uses official QA and the latter
uses QA--evidence supervision generated from the records. Evaluation covers all
1,013 official QA items, including the 600 items used for Official-QA training
and validation. ATM-Bench-Hard, the 31-question hard multi-hop subset, is reported
separately. Leaderboard references also use the full 1,013-item population.

RHELM contains 1,305 official QA items from ten personas and is evaluated on the
full official set, both with and without external data sources. The Official-QA
reference uses 524 official items for training and 264 for validation; these
items remain included in evaluation. The final MemSteer configuration uses 788
generated QA--evidence items for training, without using official QA for training
or design selection.
In the six-user setting, four personas provide training supervision,
two provide validation supervision, and the remaining four are held out
(Table~\ref{tab:rhelm-persona-breakdown}). For personal continuation, MemSteer
generates 421 supervision items from Martin Keller's history, split into 282
training and 139 validation items, and evaluates the continued design on 228
official questions. For David R.\ Ellis, it generates 337 supervision items,
split into 223 training and 114 validation items, and evaluates on 158 official
questions.

\subsection{Evaluation Metrics}
\label{app:exp-metrics}

On ATM-Bench, Question Score (QS) uses exact match
for numerical items, Jaccard partial credit for list items, and judge scores
for open-ended items. R@10 averages the fraction of gold evidence records
among the top-10 retrieved records over questions.

\noindent\textbf{Blind answer agreement.}
Blind answer recovery checks type-normalized exact agreement: normalized text
equality, set equality for lists, and exact value-and-unit agreement for
numerical answers. Boolean answers are matched exactly. Evidence coverage is
checked at the record level, requiring citations to every selected evidence record.

\noindent\textbf{Design-selection accuracy.}
Validation accuracy guides revision acceptance and design selection.
Predictions identical to their reference answers are scored as correct;
other responses receive a binary correctness judgment from a fixed LLM judge.
Final ATM-Bench evaluation uses Question Score (QS), which includes partial
credit for list questions. Final RHELM evaluation reports overall accuracy,
computed as the mean of binary correctness judgments from the official evaluator.

\subsection{Feedback inputs}
\label{app:exp-feedback-inputs}
\textit{Gold evidence} denotes the reference source evidence in
MemSteer-generated QA--evidence supervision, not human-annotated evidence.
For \textit{Random-30}, each sampled case includes the question, reference
answer, prediction and score, ranked
retrieved-record IDs, IDs actually delivered to the answer.
Gold evidence remains unmarked in Random-30, even if it appears in the context.
\textit{Random-30 + Gold evidence} adds required-evidence IDs and gold-record
excerpts without programmatic failure attribution. \textit{Random-30 + Diagnosis}
retains random sampling and additionally provides failure-stage labels,
missing-evidence IDs, evidence checks, and memory-lookup results.
\textit{Diagnosis + stratified sampling} uses the same diagnostic information
but organizes examples by failure category before allocating the feedback budget.

\FloatBarrier
\section{More Experimental Results}
\label{app:exp-more-results}

\subsection{ATM-Bench reference and optimization results}

Table~\ref{tab:atm-deepseek-appendix} reorganizes the DeepSeek-V4-Flash-evaluated
results from Table~\ref{tab:q1-atm} .

\par\addvspace{4pt}
\noindent\begin{minipage}{\linewidth}
\begingroup
\makeatletter\def\@captype{table}\makeatother
\centering
\footnotesize
\setlength{\tabcolsep}{2.2pt}
\renewcommand{\arraystretch}{1.06}
\providecommand{\gainentry}[2]{\begin{tabular}[t]{@{}c@{}}#1\\[-3pt]{\tiny\color{green!60!black}$\uparrow$#2}\end{tabular}}
\providecommand{\gainentrybold}[2]{\begin{tabular}[t]{@{}c@{}}\textbf{#1}\\[-3pt]{\tiny\color{green!60!black}$\uparrow$\textbf{#2}}\end{tabular}}
\providecommand{\lossentry}[2]{\begin{tabular}[t]{@{}c@{}}#1\\[-3pt]{\tiny\color{red!70!black}$\downarrow$#2}\end{tabular}}
\caption{Performance on ATM-Bench and ATM-Bench-Hard (\%). \textbf{DeepSeek-V4-Flash is used for answer generation and as the judge for open-ended answers.} Arrows indicate percentage-point changes relative to the corresponding initial design: Base-BM25 for MemSteer-Opt and the respective starting systems for the external optimizers. Bold and underlined values indicate the best and second-best results per metric, excluding the oracle, which uses gold evidence. $^\dagger$Official-QA training items are included in evaluation.}
\label{tab:atm-deepseek-appendix}
\begin{tabular*}{\textwidth}{@{\extracolsep{\fill}}lll cc | cc@{}}
\toprule
 & & & \multicolumn{2}{c|}{\textbf{ATM-Bench}} & \multicolumn{2}{c}{\textbf{ATM-Bench-Hard}} \\
\cmidrule(lr){4-5} \cmidrule(lr){6-7}
\textbf{Method} & \textbf{Backbone} & \textbf{Supervision} & \textbf{QS} $\uparrow$ & \textbf{R@10} $\uparrow$ & \textbf{QS} $\uparrow$ & \textbf{R@10} $\uparrow$ \\
\midrule

\rowcolor{gray!10}\multicolumn{7}{@{}l}{\textit{Reference memory systems}} \\
Memexa & DeepSeek-V4 & None & 65.3 & 78.9 & -- & -- \\
A-Mem & DeepSeek-V4 & None & -- & -- & 16.7 & 38.5 \\
MemPalace & DeepSeek-V4 & None & -- & -- & 13.6 & 28.9 \\
HippoRAG2 & DeepSeek-V4 & None & -- & -- & 12.3 & 32.2 \\
\midrule

\rowcolor{gray!10}\multicolumn{7}{@{}l}{\textit{Base starting system and oracle}} \\
Base-BM25 & DeepSeek-V4 & -- & 54.9 & 61.7 & 8.2 & 10.0 \\
Oracle & DeepSeek-V4 & -- & 71.8 & -- & 46.2 & -- \\
\midrule

\rowcolor{gray!10}\multicolumn{7}{@{}l}{\textit{MemSteer-Opt (Ours) across supervision sources}} \\
MemSteer-Opt & DeepSeek-V4 & Official-QA$^\dagger$ & \underline{67.3} & \underline{83.5} & \underline{27.9} & \underline{42.2} \\
\rowcolor{rhelmgreen}
\textbf{MemSteer-Opt} & DeepSeek-V4 & \textbf{MemSteer} & \gainentrybold{68.3}{13.4} & \gainentrybold{84.1}{22.4} & \gainentrybold{29.1}{20.9} & \gainentrybold{43.6}{33.6} \\
\midrule

\rowcolor{gray!10}\multicolumn{7}{@{}l}{\textit{External optimizers}} \\
$M^\star$ start & DeepSeek-V4 & -- & 47.4 & 55.2 & 1.9 & 13.6 \\
$M^\star$ & DeepSeek-V4 & Official-QA$^\dagger$ & 51.8 & 69.4 & 9.5 & 26.0 \\
\rowcolor{rhelmgreen}
$M^\star$ & DeepSeek-V4 & \textbf{MemSteer} & \gainentry{53.4}{6.0} & \gainentry{69.8}{14.6} & \gainentry{12.9}{11.0} & \gainentry{25.6}{12.0} \\
\addlinespace[2pt]
MemEvolve start & DeepSeek-V4 & -- & 35.1 & 41.3 & 7.1 & 18.7 \\
MemEvolve & DeepSeek-V4 & Official-QA$^\dagger$ & 53.7 & 63.8 & 11.4 & 21.6 \\
\rowcolor{rhelmgreen}
MemEvolve & DeepSeek-V4 & \textbf{MemSteer} & \gainentry{53.7}{18.6} & \gainentry{63.7}{22.4} & \gainentry{9.5}{2.3} & \lossentry{18.6}{0.1} \\
\bottomrule
\end{tabular*}
\endgroup
\end{minipage}
\par\addvspace{4pt}

For MemSteer-Opt, we retested the full QA pipeline using GPT-5-mini judge.
Table~\ref{tab:atm-combined-appendix} provides the complementary GPT-5-mini
evaluation.
Note that variability in LLM-generated retrieval queries can therefore yield
small differences in R@10 with Table~\ref{tab:atm-deepseek-appendix}.
The native reference rows come from the ATM-Bench
leaderboard~\citep{atmbench2026leaderboard}, use Qwen3-VL
backbones~\citep{bai2025qwen3vl}, and report QS from the official
GPT-5-mini~\citep{openai_gpt5mini} judge. Here, 2B/8B denotes the
memory-processing and answering model sizes, respectively.

\begin{table}[!t]
\begingroup
\makeatletter\def\@captype{table}\makeatother
\centering
\footnotesize
\setlength{\tabcolsep}{2.2pt}
\renewcommand{\arraystretch}{1.06}
\providecommand{\gainentry}[2]{\begin{tabular}[t]{@{}c@{}}#1\\[-3pt]{\tiny\color{green!60!black}$\uparrow$#2}\end{tabular}}
\providecommand{\gainentrybold}[2]{\begin{tabular}[t]{@{}c@{}}\textbf{#1}\\[-3pt]{\tiny\color{green!60!black}$\uparrow$\textbf{#2}}\end{tabular}}
\caption{Performance on ATM-Bench and ATM-Bench-Hard (\%). \textbf{MemSteer-Opt uses DeepSeek-V4-Flash for answer generation. Open-ended answers are evaluated by the official GPT-5-mini judge.} All MemSteer-Opt runs start from Base-BM25. Green arrows indicate percentage-point improvements over Base-BM25. Bold and underlined values indicate the best and second-best results per metric among the MemSteer-Opt variants, respectively. The oracle uses gold evidence. $^\dagger$Official-QA training items are included in evaluation.}
\label{tab:atm-combined-appendix}
\begin{tabular*}{\textwidth}{@{\extracolsep{\fill}}lll cc | cc@{}}
\toprule
 & & & \multicolumn{2}{c|}{\textbf{ATM-Bench}} & \multicolumn{2}{c}{\textbf{ATM-Bench-Hard}} \\
\cmidrule(lr){4-5} \cmidrule(lr){6-7}
\textbf{Method} & \textbf{Backbone} & \textbf{Supervision} & \textbf{QS} $\uparrow$ & \textbf{R@10} $\uparrow$ & \textbf{QS} $\uparrow$ & \textbf{R@10} $\uparrow$ \\
\midrule

\rowcolor{gray!10}\multicolumn{7}{@{}l}{\textit{Reference memory systems}} \\
MemPalace & Qwen3-VL-2B/8B & None & 56.8 & 76.4 & 9.7 & 28.3 \\
ATM-RAG & Qwen3-VL-2B/8B & None & 51.0 & 68.7 & 13.8 & 30.4 \\
Self-RAG & Qwen3-VL-2B/8B & None & 50.3 & 68.7 & -- & -- \\
ScrapMem & Qwen3-VL-2B/8B & None & 48.4 & 66.1 & -- & -- \\
MemoryOS & Qwen3-VL-2B/8B & None & 47.2 & 59.2 & 13.7 & 32.7 \\
A-Mem & Qwen3-VL-2B/8B & None & 44.8 & 66.4 & 9.9 & 31.7 \\
Mem0 & Qwen3-VL-2B/8B & None & 43.5 & 61.9 & 9.2 & 23.7 \\
\midrule

\rowcolor{gray!10}\multicolumn{7}{@{}l}{\textit{Base starting system}} \\
Base-BM25 & DeepSeek-V4 & -- &  55.1 & 61.7 & 5.4 & 10.0 \\
Oracle & DeepSeek-V4 & -- & 71.6 & -- & 39.5 & --  \\
\midrule

\rowcolor{gray!10}\multicolumn{7}{@{}l}{\textit{MemSteer-Opt (Ours) across supervision sources}} \\
MemSteer-Opt & DeepSeek-V4 & Naive-QA  & 54.1 & 60.6 & 5.4 & 16.7 \\
MemSteer-Opt & DeepSeek-V4 & Naive-QA + checks & \gainentry{65.4}{10.4} & \gainentry{82.0}{20.3} & \gainentry{\underline{25.6}}{20.2} & \gainentry{\underline{43.2}}{33.2} \\
MemSteer-Opt & DeepSeek-V4 & Ragas-QA & \gainentry{63.5}{8.4} & \gainentry{\underline{83.5}}{21.8} & \gainentry{23.3}{17.9} & \gainentry{42.1}{32.1} \\
MemSteer-Opt & DeepSeek-V4 & Official-QA$^\dagger$ & \gainentrybold{69.6}{14.5} & \gainentry{83.2}{21.5} & \gainentry{24.3}{18.9} & \gainentry{42.3}{32.3} \\
\rowcolor{rhelmgreen}
\textbf{MemSteer-Opt} & DeepSeek-V4 & \textbf{MemSteer} & \gainentry{\underline{69.2}}{14.2} & \gainentrybold{84.1}{22.4} & \gainentrybold{27.5}{22.1} & \gainentrybold{43.8}{33.8} \\
\bottomrule
\end{tabular*}
\endgroup
\end{table}
\par\addvspace{4pt}

\par
\subsection{Effect of supervision budget}
\label{app:atm-training-budget}

We augment the existing pool of generated QA--evidence pairs with an additional
batch of generated data. From the expanded pool, we randomly sample supervision
for five settings containing 100, 200, 300, 400, or 500 training questions,
respectively. In each setting, training and validation account for approximately
60\% and 40\% of the sampled supervision. All runs start from the same Base-BM25
design, with training questions used to guide design revisions and validation
questions used for design selection. Figure~\ref{fig:atm-training-budget} reports
the resulting performance on ATM-Bench and ATM-Bench-Hard.

\par\addvspace{4pt}
\noindent\begin{minipage}{\linewidth}
\begingroup
\makeatletter\def\@captype{figure}\makeatother
\centering
\includegraphics[width=\textwidth]{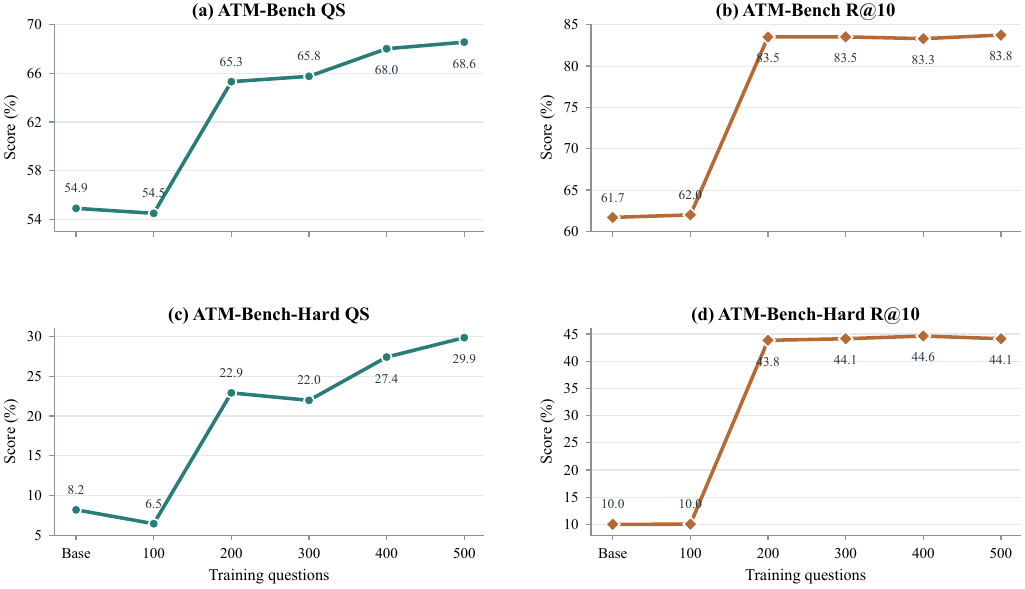}
\caption{Effect of the supervision budget on ATM-Bench and ATM-Bench-Hard (\%).
All runs start from the same Base-BM25 design, shown as Base at zero training
questions. The horizontal axis indicates the number of training questions;
the validation set scales proportionally to maintain an approximately 60:40
training-to-validation split.}
\label{fig:atm-training-budget}
\endgroup
\end{minipage}
\par\addvspace{4pt}

\noindent\begin{minipage}{\linewidth}
\begingroup
\makeatletter\def\@captype{table}\makeatother
\centering
\caption{RHELM accuracy (\%) by question type, with and without external data sources. FC, TP, AG, HL, and MI denote fact, temporal, aggregation, hallucination, and misleading questions, respectively; EX covers attachments and emails, and MX denotes mixed-context questions. Avg reports overall accuracy across evaluation questions. All memory-framework methods and our systems use GPT-4.1-mini for answer generation. \textit{ATM-Bench-evolved (frozen)} transfers the ATM-Bench-optimized memory design without further optimization on RHELM; its \textit{+ Official-QA} and \textit{+ MemSteer} variants continue optimization on RHELM using the corresponding supervision. \textit{Base-BM25+Official-QA} and \textit{Base-BM25+MemSteer} instead initialize optimization from Base-BM25. Bold and underlined values indicate the best and second-best results per column, respectively. $^\dagger$Official-QA training and validation items are included in evaluation.}\label{tab:rhelm-full}
\resizebox{\linewidth}{!}{\renewcommand{\arraystretch}{1.1}
\setlength{\tabcolsep}{4pt}
\begin{tabular}{l|ccccc|c|c|c|ccccc|c|c|c}
\toprule
\multirow{2.5}{*}{\textbf{Model}} &
\multicolumn{8}{c|}{\cellcolor{green!10}\textbf{Without External Data Sources}} &
\multicolumn{8}{c}{\cellcolor{blue!10}\textbf{With External Data Sources}}\\
\cmidrule(lr){2-9} \cmidrule(lr){10-17}
& \textbf{FC} & \textbf{TP} & \textbf{AG} & \textbf{HL} & \textbf{MI} & \textbf{EX} & \textbf{MX} & \textbf{Avg}
& \textbf{FC} & \textbf{TP} & \textbf{AG} & \textbf{HL} & \textbf{MI} & \textbf{EX} & \textbf{MX} & \textbf{Avg} \\
\midrule
\rowcolor{gray!15}\multicolumn{17}{c}{\emph{RAG Baselines}}\\
GPT-4.1-mini ($k=5$) & 35.8 & 17.3 & 17.7 & 15.2 & 3.1 & 8.0 & 10.0 & 16.3 & 34.8 & 14.1 & 19.3 & 11.7 & 3.1 & 16.9 & 12.4 & 17.5 \\
GPT-4.1-mini ($k=20$) & 44.0 & 32.4 & 31.8 & 18.3 & 3.1 & 12.1 & 12.9 & 23.5 & 42.5 & 28.7 & 30.7 & 13.2 & 3.1 & 28.5 & 13.8 & 25.1 \\
GPT-4.1-mini ($k=50$) & 59.9 & 41.6 & \textbf{40.1} & 15.7 & 1.5 & 12.9 & 16.7 & 28.9 & 54.6 & \underline{39.5} & \underline{38.0} & 11.2 & 1.5 & 38.6 & 22.4 & 32.6 \\
Hybrid ($k=5$) & 34.3 & 20.5 & 14.1 & 19.8 & 1.5 & 8.0 & 10.5 & 16.7 & 31.9 & 19.5 & 14.1 & 19.3 & 1.5 & 16.1 & 10.5 & 17.6 \\
Hybrid ($k=20$) & 47.3 & 35.7 & 31.8 & 19.3 & 3.1 & 10.4 & 15.2 & 24.8 & 45.9 & 30.8 & 26.6 & 16.8 & 4.6 & 26.9 & 15.7 & 26.0 \\
Hybrid ($k=50$) & 56.5 & 41.1 & 35.9 & 15.2 & 3.1 & \underline{13.7} & 16.7 & 27.8 & 53.1 & 37.8 & 33.9 & 8.6 & 3.1 & 33.3 & 18.6 & 29.6 \\
GPT-4.1 ($k=20$) & 51.7 & 34.1 & 35.9 & 23.9 & 7.7 & \textbf{16.1} & 17.6 & 28.2 & 50.2 & 29.2 & 32.3 & 19.8 & 6.2 & 32.5 & 19.5 & 29.5 \\
Gemini-2.5-Pro ($k=20$) & 45.4 & 35.1 & 27.1 & \underline{66.0} & \underline{23.1} & 12.4 & 18.1 & \underline{32.6} & 43.0 & 31.9 & 26.0 & \textbf{64.5} & \underline{26.2} & 31.3 & 20.5 & 35.5 \\
Claude-Opus-4.5 ($k=20$) & 50.7 & 37.8 & 33.3 & \textbf{68.0} & \textbf{47.7} & \underline{13.7} & 16.2 & \textbf{36.2} & 50.2 & 30.8 & 31.8 & \underline{60.9} & \textbf{41.5} & 33.7 & 21.0 & 38.1 \\
\rowcolor{blue!5}\multicolumn{17}{c}{\emph{Long-Context Models (1M)}}\\
Gemini-2.5-Flash-Lite & 33.2 & 22.7 & 15.2 & 17.3 & 0.0 & 9.5 & 5.6 & 16.0 & 31.7 & 14.1 & 23.4 & 7.6 & 0.0 & 19.0 & 13.1 & 17.3 \\
Qwen-2.5-14B-Instruct & 29.5 & 15.1 & 29.7 & 3.1 & 0.0 & 11.7 & 9.1 & 15.3 & 16.9 & 7.0 & 15.6 & 1.0 & 0.0 & 5.2 & 6.2 & 8.1 \\
GPT-4.1-mini & 55.1 & 31.9 & \textbf{40.1} & 4.1 & 1.5 & 11.2 & 12.4 & 24.0 & 49.3 & 27.0 & 33.9 & 2.0 & 1.5 & 43.4 & 0.3 & 33.9 \\
\rowcolor{green!5}\multicolumn{17}{c}{\emph{Memory Frameworks (GPT-4.1-mini)}}\\
MemGPT & 31.9 & 18.4 & 22.9 & 0.5 & 0.0 & 7.6 & 8.1 & 13.9 & 27.5 & 14.6 & 28.7 & 1.5 & 1.5 & 18.9 & 17.1 & 17.3 \\
Mem0 & 41.6 & 31.4 & 28.1 & 10.7 & 3.1 & 10.8 & 13.3 & 21.1 & 46.4 & 29.2 & 27.1 & 10.2 & 3.1 & 31.3 & 35.7 & 28.9 \\
MemU & 49.3 & 32.4 & 33.9 & 8.6 & 4.6 & 12.0 & 11.4 & 23.1 & 54.6 & 36.2 & 35.4 & 10.2 & 3.1 & 36.5 & 36.7 & 33.6 \\
\textcolor{rhelmlink}{Base-BM25} & 50.2 & 25.9 & 30.2 & 6.1 & 3.1 & 11.2 & 18.6 & 22.3 & 47.3 & 21.6 & 27.1 & 10.7 & 3.1 & 45.8 & 51.4 & 33.3 \\
\textcolor{rhelmlink}{Base-BM25+Official-QA$^\dagger$} & 58.5 & 26.5 & 36.5 & 19.8 & 3.1 & 9.6 & \underline{19.5} & 26.5 & 54.6 & 23.8 & 37.5 & 21.3 & 4.6 & 53.4 & \textbf{59.0} & 40.7 \\
\textcolor{rhelmlink}{Base-BM25+MemSteer} & 53.6 & 27.0 & \underline{39.1} & 16.8 & 4.6 & 12.4 & \underline{19.5} & 26.4 & 48.8 & 28.1 & 37.0 & 23.4 & 6.2 & 52.2 & \underline{55.7} & 39.9 \\
\textcolor{blue!70!black}{ATM-Bench-evolved(frozen)} & \textbf{64.3} & 29.2 & 33.3 & 9.1 & 3.1 & \underline{13.7} & 19.0 & 26.4 & \textbf{59.4} & 28.6 & 33.9 & 10.7 & 3.1 & 52.6 & 51.9 & 38.6 \\
\textcolor{blue!70!black}{ATM-Bench-evolved + Official-QA$^\dagger$} & \underline{60.4} & \textbf{47.6} & \underline{39.1} & 24.9 & 3.1 & 12.0 & \textbf{20.0} & 31.5 & \underline{56.0} & \textbf{43.2} & \textbf{39.1} & 16.2 & 4.6 & \underline{58.6} & 51.4 & \textbf{42.9} \\
\textcolor{blue!70!black}{ATM-Bench-evolved + MemSteer} & 58.5 & \underline{46.5} & \underline{39.1} & 21.8 & 3.1 & 12.0 & \textbf{20.0} & 30.6 & 55.6 & \textbf{43.2} & 35.9 & 18.8 & 3.1 & \textbf{59.0} & 51.4 & \underline{42.8} \\

\bottomrule
\end{tabular}}
\endgroup
\end{minipage}

\begin{table}[!t]
\begingroup
\makeatletter\def\@captype{table}\makeatother
\centering
\footnotesize
\setlength{\tabcolsep}{3.5pt}
\renewcommand{\arraystretch}{1.08}
\caption{Per-persona accuracy on RHELM with external data sources (\%). (a) Six-user optimization: roles indicate participation in training, validation, or neither (held-out). (b) Ten-user optimization: all personas participate as training and validation, with Official-QA or MemSteer supervision. $N$ is the number of evaluation questions; parentheses show percentage-point changes from Base-BM25. Group summaries pool evaluation questions from the six source users (training and validation) or the four held-out users; \textit{Overall (pooled)} pools all 1,305 questions. $^\dagger$Official-QA training and validation items are included in evaluation.}
\label{tab:rhelm-persona-breakdown}

\fontsize{8.5}{9.5}\selectfont
\textbf{(a) Six-user optimization setting}\par\smallskip
\begin{tabular*}{\linewidth}{@{\extracolsep{\fill}}l c r cc@{}}
\toprule
\textbf{Persona} & \textbf{Role} & \textbf{$N$} & \textbf{Base-BM25} & \textbf{Shared design} \\
\midrule
Jonas Richter & Training & 106 & 25.5 & 33.0 {\tiny\color{green!60!black}(+7.5)} \\
David Reyes & Training & 132 & 34.8 & 37.1 {\tiny\color{green!60!black}(+2.3)} \\
Dr. Elias Romero & Training & 122 & 31.1 & 31.1 {\tiny\color{gray}(+0.0)} \\
Justice Arvind P. Mehra & Training & 123 & 33.3 & 34.1 {\tiny\color{green!60!black}(+0.8)} \\
Elena Morales & Validation & 87 & 35.6 & 39.1 {\tiny\color{green!60!black}(+3.5)} \\
Dr. Elena Markovic & Validation & 109 & 28.4 & 32.1 {\tiny\color{green!60!black}(+3.7)} \\
\midrule
Dr. Alastair McEwan & Held-out & 96 & 30.2 & 34.4 {\tiny\color{green!60!black}(+4.2)} \\
Dr. Marcus Ellison & Held-out & 144 & 27.1 & 27.8 {\tiny\color{green!60!black}(+0.7)} \\
Martin Keller & Held-out & 228 & 41.2 & 37.7 {\tiny\color{red!70!black}($-$3.5)} \\
David R. Ellis & Held-out & 158 & 37.3 & 38.6 {\tiny\color{green!60!black}(+1.3)} \\
\midrule
\textit{Source users (6)} & -- & 679 & 31.5 & 34.3 {\tiny\color{green!60!black}(+2.8)} \\
\textit{Held-out users (4)} & -- & 626 & 35.3 & 35.1 {\tiny\color{red!70!black}($-$0.2)} \\
\rowcolor{gray!10}\textit{Overall (pooled)} & -- & 1,305 & 33.3 & 34.7 {\tiny\color{green!60!black}(+1.4)} \\
\bottomrule
\end{tabular*}

\par\medskip
\textbf{(b) Ten-user optimization setting}\par\smallskip
\begin{tabular*}{\linewidth}{@{\extracolsep{\fill}}l r c cc@{}}
\toprule
\textbf{Persona} & \textbf{$N$} & \textbf{Base-BM25} & \textbf{Official-QA}$^\dagger$ & \textbf{MemSteer} \\
\midrule
Jonas Richter & 106 & 25.5 & 37.7 {\tiny\color{green!60!black}(+12.2)} & 40.6 {\tiny\color{green!60!black}(+15.1)} \\
David Reyes & 132 & 34.8 & 36.4 {\tiny\color{green!60!black}(+1.6)} & 37.9 {\tiny\color{green!60!black}(+3.1)} \\
Dr. Elias Romero & 122 & 31.1 & 41.0 {\tiny\color{green!60!black}(+9.9)} & 34.4 {\tiny\color{green!60!black}(+3.3)} \\
Justice Arvind P. Mehra & 123 & 33.3 & 39.8 {\tiny\color{green!60!black}(+6.5)} & 39.8 {\tiny\color{green!60!black}(+6.5)} \\
Elena Morales & 87 & 35.6 & 47.1 {\tiny\color{green!60!black}(+11.5)} & 42.5 {\tiny\color{green!60!black}(+6.9)} \\
Dr. Elena Markovic & 109 & 28.4 & 38.5 {\tiny\color{green!60!black}(+10.1)} & 37.6 {\tiny\color{green!60!black}(+9.2)} \\
Dr. Alastair McEwan & 96 & 30.2 & 39.6 {\tiny\color{green!60!black}(+9.4)} & 37.5 {\tiny\color{green!60!black}(+7.3)} \\
Dr. Marcus Ellison & 144 & 27.1 & 41.7 {\tiny\color{green!60!black}(+14.6)} & 38.2 {\tiny\color{green!60!black}(+11.1)} \\
Martin Keller & 228 & 41.2 & 42.5 {\tiny\color{green!60!black}(+1.3)} & 43.0 {\tiny\color{green!60!black}(+1.8)} \\
David R. Ellis & 158 & 37.3 & 41.8 {\tiny\color{green!60!black}(+4.5)} & 44.3 {\tiny\color{green!60!black}(+7.0)} \\
\midrule
\rowcolor{gray!10}\textit{Overall (pooled)} & 1,305 & 33.3 & 40.7 {\tiny\color{green!60!black}(+7.4)} & 39.9 {\tiny\color{green!60!black}(+6.6)} \\
\bottomrule
\end{tabular*}
\endgroup
\end{table}

Performance remains close to Base-BM25 at the smallest question budget, with
the largest gains occurring between the 100- and 200-question settings.
Beyond this point, QS on ATM-Bench generally improves, while R@10 remains
largely stable on both benchmarks. ATM-Bench-Hard shows small fluctuations
in QS, with a test set of only 31 questions.

\par
\par\addvspace{4pt}
\subsection{Full RHELM results}
\label{app:exp-rhelm-full}

Table~\ref{tab:rhelm-full} reports RHELM results by question type, with and without
external data sources. It also includes baselines using other answer models and
long-context models for reference, rather than as controlled comparisons with
the GPT-4.1-mini systems in Table~\ref{tab:q1-rhelm}.
\par\addvspace{4pt}

\par
\par\addvspace{4pt}
\subsection{Per-persona results}

Table~\ref{tab:rhelm-persona-breakdown} reports per-persona results for shared
memory designs under two optimization settings, complementing the shared and
continued designs compared in Table~\ref{tab:q3}. In the six-user setting,
four personas provide training supervision and two provide validation
supervision; the remaining four are held out from optimization. In the ten-user
setting, all personas participate in optimization. Official-QA uses 40\% of each
persona's official QA for training and 20\% for validation, whereas MemSteer uses
QA--evidence supervision generated from user histories and reserves official QA
for evaluation.

\paragraph{Transfer of the six-user shared design.}
Table~\ref{tab:rhelm-persona-breakdown}(a) shows that the shared design improves
pooled accuracy on the six source users from 31.5\% to 34.3\%. However, this
improvement does not carry over in aggregate to the four users held out from
optimization. These results highlight heterogeneous transfer across users: a
design optimized on the source users does not necessarily benefit other users.
This makes the target-user continuation valuable, as studied in
Table~\ref{tab:q3}.

\paragraph{Generated versus official supervision in the ten-user setting.}
Table~\ref{tab:rhelm-persona-breakdown}(b) shows that MemSteer improves over
Base-BM25 for all ten personas, raising pooled accuracy from 33.3\% to 39.9\%.
Its overall accuracy is 0.8 points below the 40.7\% obtained with Official-QA
supervision. Note that MemSteer reserves official QA for evaluation, whereas the
Official-QA results include questions used for training or validation. Thus,
history-generated supervision provides an effective alternative for optimizing
a shared memory design without relying on official QA for training or design
selection.
\par\addvspace{4pt}

\par
\FloatBarrier
\subsection{Example accepted architecture trajectory}
\label{app:exp-trajectories}

\begin{minipage}{\linewidth}
\makeatletter\def\@captype{figure}\makeatother
\centering
\includegraphics[width=0.84\textwidth]{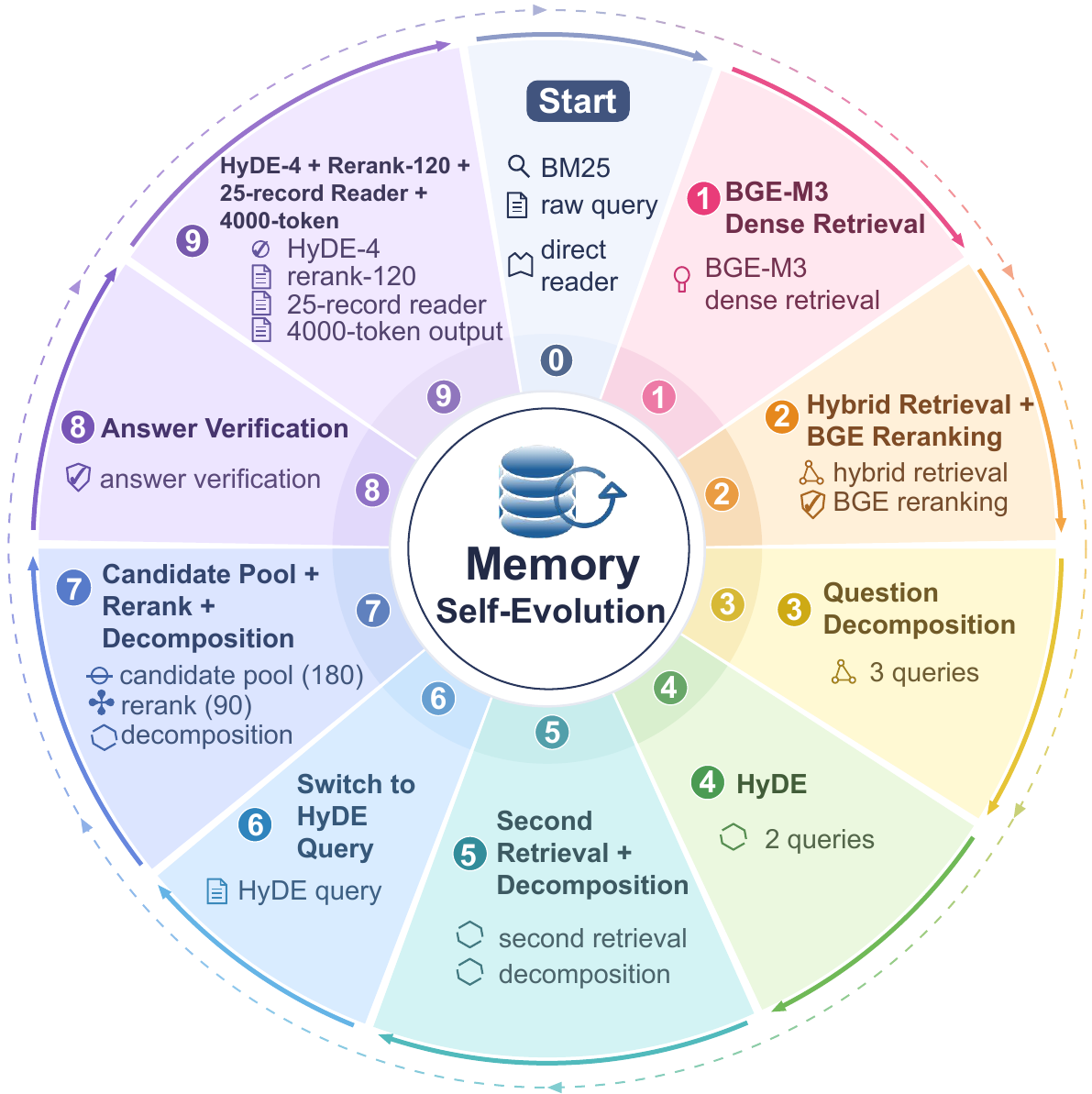}
\caption{Accepted memory-design optimization trajectory on ATM-Bench, starting
from Base-BM25 and comprising nine accepted revisions. Numbered badges and
colored arrows indicate the revision order.}
\label{fig:evolution-trajectories}
\end{minipage}

Figure~\ref{fig:evolution-trajectories} traces one accepted memory-design optimization trajectory on ATM-Bench. Starting
from Base-BM25 with raw queries and a direct reader, revisions add BGE-M3~\citep{chen2024bgem3} dense
retrieval; hybrid retrieval with BGE reranking; question decomposition; HyDE~\citep{gao2023hyde} and
a second retrieval stage; HyDE queries; a 180-candidate pool with 90 reranked
items; and answer verification. The final design uses four HyDE queries, reranks
120 candidates, reads 25 records, and permits a 4,000-token output. The sequence
moves from retrieval representation and ranking to multi-query context
construction and answer validation.

\section{Limitations}
MemSteer's supervision is generated and verified by LLMs without human
certification, so errors or biases shared across answer planning,
source-grounded auditing, and blind answer recovery can pass verification
undetected. Blind answer recovery checks whether the reference answer is
recoverable from the sampled records and every selected evidence record is
cited; it does not establish that each record is necessary.
Generated questions follow the distribution of the history rather than users'
actual information needs.
MemSteer-Opt explores a constrained, largely retrieval-centric design space.
Finally, our evaluation covers two benchmarks, one of which (RHELM) consists
of synthetic personas, so evidence on real, heterogeneous user histories
remains limited.

\par
\clearpage
\endgroup